\documentclass[natbib=False,nonacm]{acmart}
\makeatletter
\renewcommand\@formatdoi[1]{\ignorespaces}
\makeatother

\renewcommand\footnotetextcopyrightpermission[1]{} 
\usepackage[style=ACM-Reference-Format]{biblatex}
\usepackage{amsmath}
\usepackage{graphicx}
\usepackage[ruled,noend]{algorithm2e}
\usepackage{float} 
\usepackage[hypcap=false]{caption}
\usepackage{subfig}
\usepackage{enumitem}
\usepackage{color,soul}
\authorsaddresses{}
\AtBeginDocument{%
  \providecommand\BibTeX{{%
    \normalfont B\kern-0.5em{\scshape i\kern-0.25em b}\kern-0.8em\TeX}}}

\setcopyright{acmcopyright}
\copyrightyear{2021}
\acmYear{2021}

\begin{document}
\pdfpxdimen=1in
\divide\pdfpxdimen by 96 

\title{Contrastive Counterfactual Visual Explanations With Overdetermination}

\author{Adam White}
\authornote{lead authors}
\affiliation{%
  \institution{City Data Science Institute - City, University of London}
  \streetaddress{}
  \city{London}
  \country{UK}
  \postcode{EC1V 0HB}}
\email{adam.whitet@city.ac.uk}

\author{Kwun Ho Ngan}
\authornotemark[1]
\affiliation{%
  \institution{City Data Science Institute - City, University of London}
  \streetaddress{}
  \city{London}
  \country{UK}
  \postcode{EC1V 0HB}}
\email{kwun-ho.ngan@city.ac.uk}

\author{James Phelan}
\affiliation{%
  \institution{City Data Science Institute - City, University of London}
  \streetaddress{}
  \city{London}
  \country{UK}
  \postcode{EC1V 0HB}}
\email{James.Phelan@city.ac.uk}

\author{Saman Sadeghi Afgeh}
\affiliation{%
  \institution{City Data Science Institute - City, University of London}
  \streetaddress{}
  \city{London}
  \country{UK}
  \postcode{EC1V 0HB}}
\email{saman.sadeghiafgeh@gmail.com}

\author{Kevin Ryan}
\affiliation{%
  \institution{City Data Science Institute - City, University of London}
  \streetaddress{}
  \city{London}
  \country{UK}
  \postcode{EC1V 0HB}}
\email{kevin.ryan.2@city.ac.uk}

\author{Constantino Carlos Reyes-Aldasoro}
\affiliation{%
  \institution{City Data Science Institute - City, University of London}
  \streetaddress{}
  \city{London}
  \country{UK}
  \postcode{EC1V 0HB}}
\email{Constantino-Carlos.Reyes-Aldasoro@city.ac.u}

\author{Artur d'Avila Garcez}
\affiliation{%
  \institution{City Data Science Institute - City, University of London}
  \streetaddress{}
  \city{London}
  \country{UK}
  \postcode{EC1V 0HB}}
\email{A.Garcez@city.ac.uk}

\renewcommand{\shortauthors}{White and Ngan, et al.}

\begin{abstract}
A novel explainable AI method called CLEAR \textit{Image} is introduced in this paper. CLEAR \textit{Image} is based on the view that a satisfactory explanation should be contrastive, counterfactual and measurable. CLEAR \textit{Image} explains an image’s classification probability by contrasting the image with a corresponding image generated automatically via  adversarial learning. This enables both salient segmentation and perturbations that faithfully determine each segment’s importance. CLEAR \textit{Image} uses regression to determine a causal equation describing a classifier's local input-output behaviour. Counterfactuals are also  identified, that are supported by the causal equation. Finally, CLEAR  \textit{Image} measures the fidelity of its explanation against the classifier. CLEAR \textit{Image} was successfully applied to a medical imaging case study where it outperformed methods such as Grad-CAM and LIME by an average of 27\% using a novel pointing game metric. CLEAR \textit{Image} excels in identifying cases of ‘causal overdetermination’ where there are multiple patches in an image, any one of which is sufficient by itself to cause the classification probability to be close to one.

\end{abstract}



\maketitle
\thispagestyle{empty}

\section{Introduction}
Data-driven AI for Computer Vision can achieve high levels of predictive accuracy, yet the rationale behind these predictions is often opaque. This paper proposes a novel explainable AI (XAI) method called CLEAR \textit{Image} that seeks to reveal the causal structure implicitly modelled by an AI system, where the causes are an image’s segments and the effect is the AI system’s classification probability. The explanations are for single predictions and describe the local input-output behaviour of the classifier. CLEAR \textit{Image} is based on the philosopher James Woodward’s seminal analysis of causal explanation \cite{Woodward}, which develops Judea Pearl’s manipulationist account of causation \cite{Pearl}. Together they constitute the dominant accounts of explanation in the philosophy of science. We argue that a successful explanation for an AI system should be contrastive, counterfactual and measurable.

According to Woodward, to explain an event \textit{E} is “to provide information about the factors on which it depends and exhibit how it depends on those factors”. This requires a \emph{causal equation} to describe the causal structure responsible for generating the event. The causal equation must support a set of counterfactuals; a counterfactual specifies a possible world where, contrary to the facts, a desired outcome occurs. The counterfactuals serve to illustrate the causal structure, and to answer a set of ‘what-if-things-had-been-different’ questions. In XAI, counterfactuals usually state  minimal changes needed to achieve the desired outcome.

A contrastive explanation seeks to answer the question \textit{‘Why E rather than F?’} \textit{F} comes from a contrast class of events that were alternatives to \textit{E}, but which did not happen \cite{van1980scientific}. An explanation identifies the causes that led to \textit{E} occurring rather than \textit{F}, even though the relevant contrast class to which \textit{F} belongs is often not explicitly conveyed.

For Woodward, all causal claims are counterfactual and contrastive: ‘to causally explain an outcome is always to explain why it, rather than some alternative, occurred’. Woodward's theory of explanation stands in opposition to the multiple XAI methods that claim to provide counterfactual explanations \cite{verma2020counterfactual}, but which only provide statements of single or multiple counterfactuals. As this paper will illustrate, counterfactuals will only provide incomplete explanations without a supporting causal equation.

CLEAR \textit{Image} excels at identifying cases of 'causal overdetermination'. The causal overdetermination of an event occurs when two or more sufficient causes of that event occur. A standard example from the philosophy literature is of soldiers in a firing squad simultaneously shooting a prisoner, with each shot being sufficient to kill the prisoner. The death of the prisoner is causally overdetermined. This causal structure may well be ubiquitous in learning systems. For example, there may be multiple patches in a medical image, any of which being sufficient by itself to cause a classification probability close to one. To the best of our knowledge, CLEAR \textit{Image} is the first XAI method capable of identifying causal overdetermination. 

 CLEAR \textit{Image} explains an image’s classification probability by contrasting the image with a corresponding GAN generated image. Previously, XAI use of GANs has just focused on their difference masks, which are created by subtracting the original image from its corresponding GAN generated image. However, as we will illustrate, a difference mask should only be a starting point for segmentation and explanation. This is because the segments identified from a difference mask can vary significantly in their relevance to a classification; furthermore, other segments critical to the classification can often be absent from the mask. Therefore, CLEAR \textit{Image} uses a novel segmentation method that combines information from the difference mask, the original image and the classifier's behaviour. After completing its segmentation, CLEAR \textit{Image} identifies counterfactuals and then follows a process of perturbation, whereby segments of the original image are changed, and the change in outcome is observed to produce a regression equation. The regression equation is used to determine the contribution each segment makes to the classification probability. As we will show, the explanations provided by leading XAI methods such as LIME and Grad-CAM often cannot be trusted. CLEAR  \textit{Image}, therefore, measures the fidelity of its explanation against the classifier, where fidelity refers to how closely an XAI method is able to mimic a classifier's behaviour.

CLEAR \textit{Image} was evaluated in two case studies, both involving overdetermination. The first uses a multifaceted synthetic dataset, and the second uses chest X-rays. CLEAR \textit{Image} outperformed XAI methods such as LIME and Grad-CAM by an average of 31\% on the synthetic dataset and 27\% on the X-ray dataset based on a pointing game metric defined in this paper for the case of multiple targets. Our code will be made available on GitHub.

The contribution of this paper is four-fold. We introduce an XAI method that: 
\begin{itemize} [topsep=0pt,itemsep=-1ex,partopsep=1ex,parsep=1ex]
    \item generates contrastive, counterfactual and measurable explanations outperforming established XAI methods in challenging image domains;
    \item uses a GAN-generated contrast image in determining a causal equation, segment importance scores and counterfactuals.
    \item offers novel segmentation and pointing game algorithms for the evaluation of image explanations.
    \item is capable of identifying causal overdetermination, i.e. the multiple sufficient causes for an image classification.

\end{itemize}

CLEAR \textit{Image} is a substantial development of an earlier XAI method, (\textbf{C}ounterfactual \textbf{L}ocal \textbf{E}xplanations vi\textbf{A} \textbf{R}egression), which only applies to tabular data \cite{white2019measurable}. New functionality includes: (i) the novel segmentation algorithm (ii) generating perturbed images by infilling from the corresponding GAN image (iii) a novel pointing game suitable for images with multiple targets (iv) identification of sufficient causes and overdetermination (v) measurement of fidelity errors for counterfactuals involving categorical features.

The remainder of the paper is organised as follows: Section 2 defines the relevant notation and background. Section 3 discusses the immediate related work. Section 4 introduces the CLEAR \textit{Image} method and algorithms. Section 5 details the experimental setup and discusses the results. Section 6 concludes the paper and indicates directions for future work.

\section{Background}

\subsection{Key Notation}
This paper adopts the following notation: let the instance $x$ be an image, and $m$ be a machine learning system that maps $x$ to class label $l$ with probability $y$. Let $x$ be partitioned into $S$ segments (regions) $\{s_1, \dots ,s_n\}$. Let any variable with a prime subscript $'$ be the variable from the GAN-generated image, e.g. $x'$ is the GAN generated image derived from $x$, and maps to class $l$ with probability $y'$. 

\subsection{Explanation by Perturbation}
Methods such as Occlusion \cite{zhou2016learning}, Extremal Perturbation \cite{fong2019understanding}, FIDO \cite{chang2018explaining}, LIME \cite{LIME} and Kernel SHAP \cite{lundberg2017unified} use perturbation to evaluate which segments of an image $x$ are most responsible for $x$’s classification probability \textit{y}. The underlying idea is that the contribution that a segment $s_i$ makes to $y$ can be determined by substituting it with an uninformative segment $s_i'$, where  $s_i'$ may be either grey, black or blurred \cite{zhou2016learning,fong2019understanding,LIME} or in-painted without regard to any contrast class \cite{chang2018explaining}. There are three key problems with using perturbed images to explain a classification:

\begin{enumerate}[topsep=0pt,itemsep=-1ex,partopsep=1ex,parsep=1ex]

\item A satisfactory explanation must be contrastive; it must answer \textit{‘Why E rather than F?’} None of the above methods does this. Their contrasts are instead images of uninformative segments. 

\item The substitution may fail to identify the contribution that $s_i$ makes to $y$. For example, replacing $s_i$ with black pixels can take the entire image beyond the classifier’s training distribution. By contrast, blurring or uninformative in-painting might result in $s_i'$ being too similar to $s_i$ resulting in the contribution of $s_i$ being underestimated.

\item A segmentation needs to be relevant to its explanatory question. Current XAI perturbation approaches produce radically different segmentations. FIDO and Extremal Perturbation identify ‘optimal’ segments that, when substituted by an uninformative segment, maximally affect the classification probability; by contrast, LIME uses a texture/intensity/colour algorithm (e.g. Quickshift \cite{vedaldi2008quick} ).

\end{enumerate}

CLEAR \textit{Image} uses GAN generated images to address each of these problems: (i) its foil is a GAN generated image $x'$ belonging to a contrast class selected by the user. (ii) inpainting with segments derived from $x'$ enables better estimation of each segment's contribution to the difference between probabilities $y$ and $y'$. (iii) the differences between $x$ and $x'$ are used to guide the segmentation.\\ 

\section{Related Work}

The XAI methods most relevant to this paper can be broadly grouped into four types:

\textbf{(i) Counterfactual methods:} Wachter et al. \cite{wachter2017counterfactual} first proposed using counterfactuals as explanations of single machine learning predictions. Many XAI methods have attempted to generate ‘optimal’ counterfactuals; for example, \cite{karimi2020survey} review sixty counterfactual methods. The algorithms differ in their constraints and the attributes referenced in their loss functions \cite{verma2020counterfactual}. Desiderata often include that a counterfactual is: (1) actionable – e.g. does not recommend that a person reduces their age, (2) near to the original observation - common measures include Manhattan distance, L1 norm and L2 norm, (3) sparse – only changing the values of a small number of features, (4) plausible - e.g. the counterfactual must correspond to a high density part of the training data, (5) efficient to compute. Karimi et al. \cite{karimi2021algorithmic} argue that these methods are likely to identify counterfactuals that are either suboptimal or infeasible in terms of their actionability. This is because they do not take account of the causal structure that determines the consequences of the person’s actions. The underlying problem is that unless all of the person’s features are causally independent of each other, then when the person acts to change the value of one feature, other downstream dependent features may also change. However, this criticism does not apply to CLEAR \textit{Image}. CLEAR \textit{Image}'s purpose is to explain the local input-output behaviour of a classifier, and the role of its counterfactuals is (i)  to illustrate the classifier's causal structure (at the level of how much each segment can cause the classification probability to change) and (ii) to answer contrastive questions. Hence if the explanatory question is "why is this image classified as showing a zebra and not a horse?", CLEAR  \textit{Image} might highlight the stripes on the animal as being a cause of the classification. Whilst this might be a satisfactory explanation of the classification, it is, of course, not actionable. In this paper, we provide a different criticism of counterfactual methods: that they fail to provide satisfactory explanations because they do not provide a causal equation describing the local behaviour of the classifier they are meant to explain. Without this, they cannot identify: the relative importance of different features, how the features are taken to interact with each other, or the functional forms that the classifier is, in effect, applying to each feature. They will also fail to identify cases of overdetermination.\\
\textbf{(ii) Gradient-based methods:} These provide saliency maps by backpropagating an error signal from a neural network’s output to either the input image or an intermediate layer. \citeauthor{simonyan2014deep} \shortcite{simonyan2014deep} use the derivative of a class score for the image to assign an importance score to each pixel. \citeauthor{kumar2017explaining}'s  \textbf{CL}ass-\textbf{E}nhanced \textbf{A}ttention \textbf{R}esponse \cite{kumar2017explaining} uses backpropagation to visualise the most dominant classes; this should not be confused with our method. A second approach modifies the backpropagation algorithm to produce sharper saliency maps, e.g. by suppressing the negative flow of gradients. Prominent examples of this approach \cite{springenberg2014striving,zeiler2014visualizing} have been found to be invariant to network re-parameterisation or the class predicted \cite{adebayo2018sanity, nie2018theoretical}. A third approach \cite{selvaraju2017grad, chattopadhay2018grad} uses the product of gradients and activations starting from a late layer. In Grad-CAM \cite{selvaraju2017grad}, the product is clamped to only highlight positive influences on class scores.\\  
\textbf{(iii) Perturbation based methods:} LIME and Kernel SHAP generate a dataset of perturbed images, which feeds into a regression model, which then calculates segment importance scores (LIME) or Shapley Values (Kernel SHAP). These bear some similarity to CLEAR \textit{Image} but key differences include: they do not use a GAN generated image, do not identify counterfactuals and do not report fidelity. 
Extremal Perturbation uses gradient descent to determine an optimal perturbed version of an image that, for a fixed area, has the maximal effect on a network’s output whilst guaranteeing that the selected segments are smooth. FIDO uses variational Bernoulli drop to find a minimal set of segments that would change an image's class. In contrast to LIME, Kernel SHAP and Extremal Perturbation, FIDO uses a GAN to in-paint segments with 'plausible alternative values'; however, these values are not generated to belong to a chosen contrast class. Furthermore, segment importance scores are not produced.\\
\textbf{(iv) GAN difference methods:}  Generative adversarial network (GAN) \cite{goodfellow2014generative} has been widely applied for synthetic image generation. Image translation through direct mapping of the original image to its target class has gained popularity, such as CycleGAN \cite{zhu2017unpaired} and StarGAN \cite{choi2018stargan}.
StarGan V2 \cite{choi2020starganv2} introduced a style vector for conditional image translation and produced high quality images over a diverse set of target conditions. These models, however, do not keep the translation minimal and make modification even for intra-domain translation. Fixed-point GAN \cite{siddiquee2019learning} introduced an identity loss penalising any deviation of the image during intra-domain translation. This aims to enhance visual similarity with the original image. DeScarGAN \cite{wolleb2020descargan} adopts this loss function in its own GAN architecture and has outperformed Fixed-point GAN in their case study for Chest X-Ray pathology identification and localisation.\\

CLEAR \textit{Image} builds on the strengths of the above XAI methods but also addresses key shortcomings. As already outlined, it uses a 'GAN-augmented' segmentation algorithm rather than just a difference mask. Next, methods such as LIME and Kernel SHAP assume that a classification probability is a simple linear addition of its causes. This is incorrect for cases of causal overdetermination, and CLEAR \textit{Image}, therefore, uses a sigmoid function (see section 4.2).  Finally, our experiments confirm that prominent XAI methods often fail to identify the most relevant regions of an image; CLEAR \textit{Image}, therefore, measures the fidelity of its explanations.

\section{The CLEAR \textit{Image} Method}
\begin{figure}[!hb]\setlength\belowcaptionskip{-18pt}
  \includegraphics[width=0.9\textwidth]{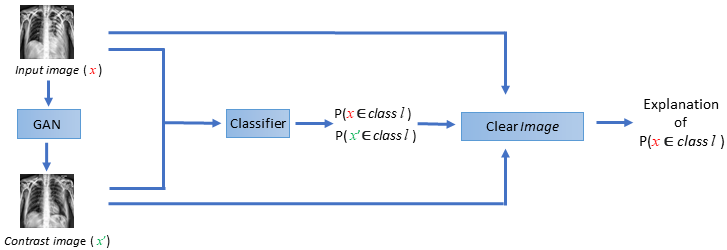}  
  \centering
  \caption{The CLEAR $Image$ pipeline. The GAN produces a contrast image. CLEAR \textit{Image} explains the classification probability by comparing the input image with its contrast image. It produces a regression equation that measures segment scores, reports fidelity and identifies cases of overdetermination. In this example, class $l$ is 'pleural effusion' and its contrast class $l'$ is 'healthy'. Using our Densenet model, the X-ray shown in this figure had a probability of belonging to $l$ equal to 1, and its contrast image had a probability of belonging to $l$ equal to 0.} 
  \Description{An overview of the CLEAR $Image$ pipeline for model explanation.}
  \label{pipeline}
\end{figure}

CLEAR \textit{Image} is a model-agnostic XAI method that explains the classification of an image made by any classifier (see Figure \ref{pipeline}). It requires both an image $x$ and a contrast image $x'$ generated by a GAN. CLEAR \textit{Image}  segments $x$ into $\{s_1, \dots ,s_n\} \in$ $S$ and then applies the same segmentation to $x'$ creating $\{s_1', \dots ,s_n'\} \in S'$.  It then determines the contributions that different subsets of \textit{S} make to \textit{y} by substituting with the corresponding segments of $S'$. CLEAR \textit{Image} is GAN agnostic, allowing the user to choose the GAN architecture most suitable to their project. A set of ‘image-counterfactuals’ $\{c_1 \dots c_k\}$ is also identified. Figures \ref{pipeline} to \ref{CLEAR_report} provide a running example of the CLEAR \textit{Image} pipeline, using the same X-ray taken from the CheXpert dataset.
\setlength\intextsep{10pt}

\subsection{GAN-Based Image Generation}
\label{sec:GAN-Based_Image_Generation}
\begin{figure}[!htbp]
\setlength\belowcaptionskip{-24pt}
  \centering
  \includegraphics[width = 0.8\textwidth]{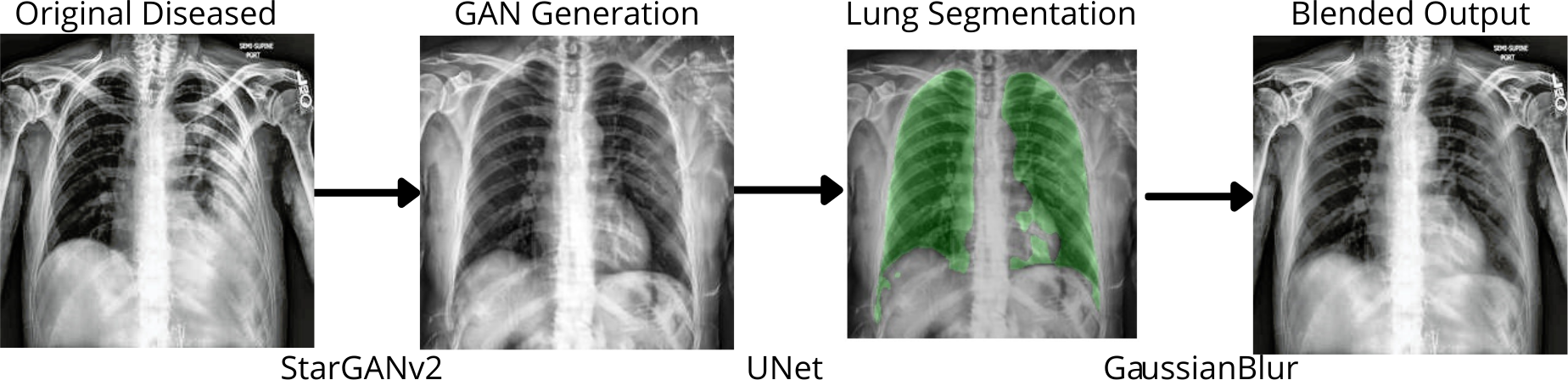}  
  \caption{The process of generating a contrast image. An original diseased image is firstly used to generate a healthy contrast image with a trained GAN model. In this example, StarGAN v2 is used as the architecture. The generated healthy lung airspace is then segmented using a U-Net segmentation model blended onto the original diseased image to produce the final image by applying GaussianBlur to minimise any edging effect around the segments.}
  \Description{The process of generating a contrast image through GAN, Segmentation and Gaussian Blur.}
  \label{postGANpipeline}
\end{figure}

To generate contrastive images, StarGAN-V2 \cite{choi2020starganv2}  and DeScarGAN \cite{wolleb2020descargan} have been deployed as the network architectures for our two case studies, the first using CheXpert, the second using a synthetic dataset. The use of these established GAN networks demonstrates how the generated contrastive images can aid in the overall CLEAR \textit{Image} pipeline. Default training hyperparameters were applied unless otherwise stated. Details of model training and hyperparameters can be found in Appendix B. The source image was used as input for the Style Encoder instead of a specific reference image for StarGAN-V2. This ensures the generated style mimics that of the input source images. StarGAN-V2 is also not locally constrained (i.e. the network will modify all pixels in an image related to the targeted class, which will include irrelevant spurious regions of the image). A post-generation lung space segmentation step using a pre-trained U-Net model  \cite{ronneberger2015unet} was therefore implemented. The original diseased lung space was replaced with the generated image, with a Gaussian Blur process to fuse the edge effect (see Figure \ref{postGANpipeline}). This confines the feature identification space used by CLEAR \textit{Image} to the lung space. It is an advantage of the CLEAR \textit{Image} pipeline that it is possible to use pre-processing to focus the explanation on the relevant parts of $x$. As we will show, XAI methods that do not take a contrast image as part of their input can sometimes identify parts of $x$, known to be irrelevant, as being responsible for the classification. 
\setlength\intextsep{15pt}

\subsection{Generating Contrastive Counterfactual Explanations}

\begin{definition}
An \textbf{image-counterfactual} $c_j$ from \textit{l} to $l'$ is an image resulting from a change in the values of one or more segments $S$ of \textit{x} to their corresponding values in
$S'$ such that class$(m(x)) = l$, class$(m(c_j)) = l'$ and $l \neq l'$. The change is minimal in that if any of the changed segments had remained at its original value, then class$(m(x))$ = class$(m(c_j))$.
\end{definition}

CLEAR \textit{Image} uses a regression equation to quantify the contribution that the individual segments make to $y$. It then measures the fidelity of its regression by comparing the classification probability resulting from each $c_j$ with an estimate obtained from the regression equation.

\begin{definition} \textbf{Counterfactual-regression fidelity error} Let $reg(c_j)$ denote the application of the CLEAR \textit{Image} regression equation given image-counterfactual $c_j$.
\begin{center}
Counterfactual-regression fidelity error $= | reg(c_j) - y_{c_j} |$.
\end{center}
\end{definition}


The following steps generate an explanation of prediction $y$ for image $x$:

\begin{enumerate} [topsep=0pt,itemsep=-1ex,partopsep=1ex,parsep=1ex]

\item GAN-Augmented segmentation algorithm. This algorithm is based on our findings (in Section 5.4) that the segments ($S_h$) determined by analysing high intensity differences between an image $x$ and its corresponding GAN generated image $x'$ will often miss regions of $x$ that are important to explaining $x$’s classification. It is therefore necessary to supplement segments $S_h$ with a second set of segments $S_l$ confined to those regions of $x$ corresponding to low intensity differences between $x$ and $x'$. $S_l$ is created based on similar textures/intensities/colours solely within $x$.\\

Pseudocode for our algorithm is shown in Algorithm \ref{Gan_seg_algo}. First, high and low thresholds ($T_h$ and $T_l$) are determined by comparing the differences between $x$ and $x'$ using multi-Otsu; alternatively the thresholds can be user-specified. $T_h$ is then used to generate a set of segments, $S_h$. The supplementary segments $S_l$, are determined by applying the low threshold, $T_l$, to the low intensity regions and then applying a sequence of connected component labelling, erosion and Felzenszwalb\cite{felzenszwalb2004efficient}. The combined set of segments, {$S_h$ and $S_l$}, is checked to see if any individual segment is an image-counterfactual. If none is found, an iterative process is applied to gradually increase the minimum segment size parameter. The final set of segments (S, S’) is subsequently created using the combined set ($S_h$, $S_l$) as shown in Figure \ref{Segmentation step}.\\

\begin{figure}[!htbp]
  \includegraphics[width=0.8\textwidth]{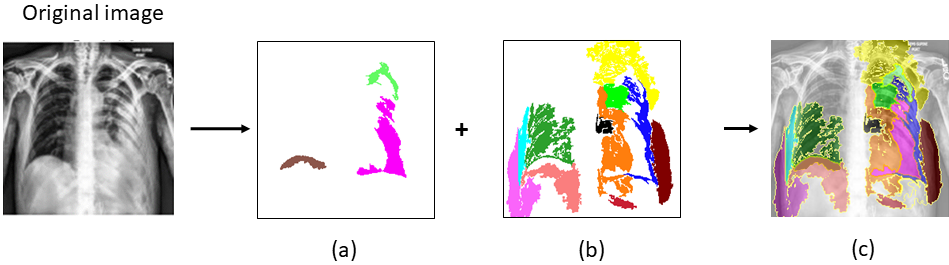}  
  \centering
  \caption{The GAN-Augmented segmentation algorithm. There are three stages. First, segments are identified from the high intensity differences between the original image $x$ and its contrast image $x'$ (a). Second, additional segments are identified from the regions of $x$ corresponding to low intensity differences between $x$ and $x'$ (b) Third, the segments from the two steps are combined (c).}
  \Description{An illustration of the GAN-Augmented segmentation algorithm.}
  \label{Segmentation step}
\end{figure} 

    
    
    


\begin{figure}[!htbp]
  \centering
    \begin{minipage}[h]{.8\textwidth}
    \begin{algorithm}[H]
    \SetKwInOut{Input}{input}\SetKwInOut{Output}{output}
    \SetKwInOut{Parameter}{Segmentation Parameters}
    \SetAlgoLined
    \DontPrintSemicolon
    \caption{GAN\_Augmented\_Segmentation}
    \Input{ $x$ - diseased image, $x'$ - contrast image, $m$ - AI classifier}
    \BlankLine
    \Parameter{
    $min\_num\_S_l$ - min number of $S_l$ segments,\\
    
    $min\_seg\_size$ - min segment size,\\ $seg\_size\_increment$ - segment size increment \newline}
    ${T_h, T_l}\leftarrow\ \;$Determine\_Thresholds({$x, x'$})\;
    ${D_h, D_l}\leftarrow\ \;$Create\_Difference\_masks({$x, x', T_h, T_l$})\;
    ${S_h}\leftarrow\ \;$Create\_high\_intensity\_segments({$D_h, min\_seg\_size$})\;
  \tcp{Connected components, erosion and Felzenswalb are now used to create the low intensity segments}
    ${S_l}\leftarrow\ \;$Create\_low\_intensity\_segments({$D_l, x, min\_seg\_size$})\;
    $N_l\leftarrow\ \;Count\_S_l\_Segments({S_l})$\;
     $N_c\leftarrow\ \;Count\_Single\_Segment\_Counterfactuals$({$m, S_h, S_l$})\;
    \tcp{If there are no counterfactuals then increase the size of the $S_l$ segments}
    \While{$N_c = 0$ and $min\_num\_S_l < N_l$}{
    $min\_seg\_size$ += $seg\_size\_increment$\;
    ${S_l}\leftarrow\ \;$Create\_low\_intensity\_segments({$D_l, x, min\_seg\_size$})\;
    $N_l\leftarrow\ \;Count\_S_l\_Segments({S_l})$\;
    
    $N_c\leftarrow\ \;Count\_Single\_Segment\_Counterfactuals$({$m, S_h, S_l$})\;}

    $S,S'\leftarrow\ \;$Add\_Segments({$S_h, S_l, x, x'$})\;    
    \Return  $S,S'$
    \label{Gan_seg_algo}
    \end{algorithm}
    \end{minipage}
\end{figure}

\item Determine $x$’s image-counterfactuals. A dataset of perturbed images is created by selectively replacing segments of $x$ with the corresponding segments of $x'$ (see Figure \ref{Counterfactual step}). A separate image is created for every combination in which either 1, 2, 3, or 4 segments are replaced. Each perturbed image is then passed through $m$ to determine its classification probability. All image-counterfactuals involving changes in up to four segments are then identified. (The maximum number of perturbed segments in a counterfactual is a user parameter; the decision to set it to 4 in our experiments was made as we found counterfactuals involving 5+ segments to have little additional explanatory value.)

\begin{figure}[htbp]
  \includegraphics[width=0.9\textwidth]{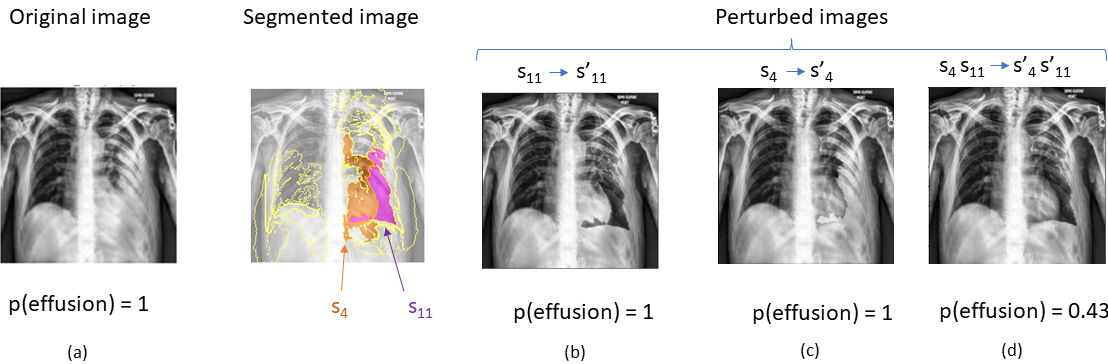}  
  \centering
  \caption{Determining image-counterfactuals. In this example segments $s_4$ and $s_{11}$ are evaluated both separately and in combination. Substituting $s_{11}$ with its corresponding contrast segment $s'_{11}$ creates a perturbed image (b) with the same classification probability as the original image (a). The same applies with segment $s_4$ (c). However substituting both segments $s_4$ and $s_{11}$ results in a perturbed image (d) which has a classification probability of 0.43. Given a decision boundary at probability of 0.5, (d) would be classified as a 'healthy' X-ray and would therefore be an image-counterfactual.}
  \Description{An illustration of determining image-counterfactual through perturbation of image segments.}
  \label{Counterfactual step}
\end{figure} 


\item Perform a stepwise logistic regression. A tabular dataset is created by using a \{0,1\} representation of the segments in each perturbed image from step 2. Consider a perturbed image $x_{per}$. This will be composed of a combination of segments $s_i$ from the original image $x$ and segments $s'_i$ from the GAN contrast image $x'$. In order to represent $x_{per}$ in tabular form, each segment of $x_{per}$ that is from $x$ is represented as a 1 and each segment of $x_{per}$ that is from $x'$ is represented as a 0. For example, if $x_{per}$ consisted solely of $\{s'_1,s_2,s_3,s_4\}$, and had a classification probability from $m$ equal to 0.75 of being 'pleural effusion', then this would be represented in tabular form as $\{0,1,1,1,0.75\}$. The table of representation vectors can then be used to generate a weighted logistic regression in which those perturbed images that are image-counterfactuals are given a high weighting and act as soft constraints.

\item Calculate segment importance scores. These are the regression coefficients for each segment from step 3.
\item Identify cases of causal overdetermination (see below).
\item Measure the fidelity of the regression by calculating fidelity errors (see Figure \ref{CLEAR_report}) and goodness of fit statistics.
\item Iterate to the best explanation. Because CLEAR \textit{Image} produces fidelity statistics, its parameters can be changed to achieve a better trade-off between interpretability and fidelity. For example, increasing the number of segments in the regression equation and including interaction terms might each increase the fidelity of an explanation but reduce its interpretability.

\end{enumerate}

\begin{figure}[!htbp]
  \includegraphics[width=0.9\textwidth]{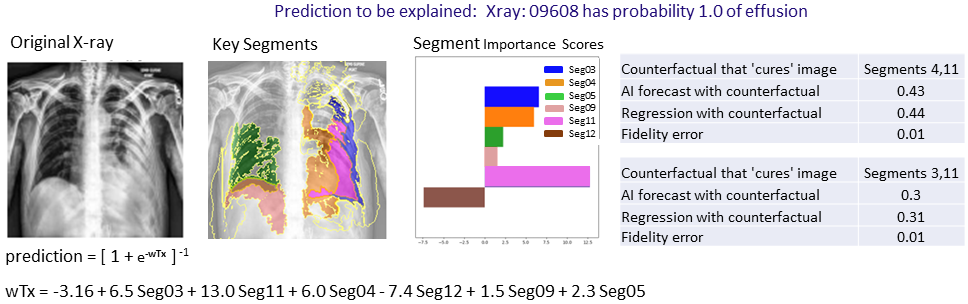}  
  \centering
  \caption{Extracts from a CLEAR \textit{Image} report. The report identifies that substituting both segments 4 and 11 with the corresponding segments from its contrast image flips the classification probability to 'healthy' According to the logistic regression equation these substitutions would change the probability of the X-ray being classified as 'pleural effusion' to 0.44. However, when these segments are actually substituted and passed through the classifier, the probability changed to 0.43, hence the fidelity error is 0.01. CLEAR \textit{Image} also identifies that substituting segments 3 and 11 also creates an image-counterfactual. Note that unlike methods such as GradCAM, CLEAR \textit{Image} is able to identify segments that have a negative impact on a classification probability.}
  \Description{An extract from a CLEAR $Image$ report.}
  \label{CLEAR_report}
\end{figure}

For CLEAR \textit{Image} an explanation is a tuple $< G;C; r; O, e >$, where $G$ are segment importance scores, $C$ are image-counterfactuals, $r$ is a regression equation, $O$ are the causes resulting in overdetermination, and $e$ are fidelity errors. The regression equation is a causal equation with each independent variable (each referring to whether a particular segment is from $x$ or $x'$) being a \textit{direct cause} of the classification probability. Figure \ref{CLEAR_report} shows an extract from a CLEAR report. Pseudocode summarising how CLEAR  \textit{Image} generates an explanations is provided in Algorithm 2.

\begin{figure}[!hbp]
  \centering
    \begin{minipage}[h]{.7\textwidth}
    \begin{algorithm}[H]
    \SetKwInOut{Input}{input}\SetKwInOut{Output}{output}
    \SetAlgoLined
    \DontPrintSemicolon
    \caption{CLEAR \textit{Image}}
    \Input{ $x$ - input image,\\ $x'$ - contrast image,\\ $m$ - AI classifier.}
    $S,S'\leftarrow\ \;$GAN\_Augmented\_Segmentation({$x,x',m$})\\ 
    \textit{D}$\leftarrow\ \;$Create\_Perturbed\_Data({$S,S',m$})\\ 
    $C \leftarrow\ \;$Find\_Counterfactuals(\textbf{$S, S', m$}) \\
    $r \leftarrow $Find\_Regression\_Equation$(D,C)$\\
    $G \leftarrow $Extract\_Segment\_Scores$(r)$\\
    $O \leftarrow $Find\_Overdetermination$(r)$\\
    $e \leftarrow  \;$Calculate\_Fidelity($C,r$)\\
    \Return  explanation$=<G,C,r,O,e>$
    \label{clear_algo}
    \end{algorithm} 
    \end{minipage}
\end{figure}

\begin{figure}[!t]
  \includegraphics[width=0.9\textwidth]{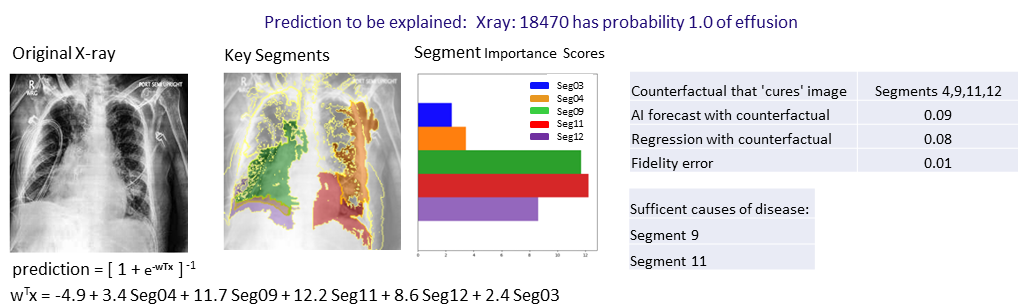}  
  \centering
  \caption{Overdetermination. The report identifies segments 9 and 11 as each sufficient to have caused the original X-ray to be classified as 'pleural effusion' with a probability greater than 0.99. Hence this is a case of causal overdetermination. The corresponding GAN-generated image $x'$ has a classification probability $\approx 0$ for pleural effusion. If a perturbed image $x_{per}$ was created by substituting all the segments of the original image $x$ with the corresponding segments of $x'$ \textit{except} for segment 9, then $x_{per}$ would  still have a classification probability for pleural effusion greater than 0.99. The same would apply if only segment 11 was substituted.}
  \Description{An illustration of reporting overdetermination from CLEAR $Image$.}
  \label{CLEAR_over}
\end{figure}

The causal overdetermination of an effect occurs when multiple sufficient causes of that effect occur. By default, CLEAR \textit{Image} only reports sufficient causes which each consist of a single segment belonging to $S$.  Substituting a sufficient cause for its corresponding member in $S'$ guarantees the effect. In the philosophy of science, it is generally taken that for an effect to be classified as overdetermined, it should be narrowly defined, such that all the sufficient causes have the same, or very nearly the same impact \cite{paul2009counterfactual}. Hence for the case studies, the effect is defined as $p(x \in diseased)> 0.99$, though the user may choose a different probability threshold. A sufficient cause changes a GAN generated healthy image to a diseased image. This is in the opposite direction to CLEAR \textit{Image}’s counterfactuals whose perturbed segments flip the classification to 'healthy'. Sufficient causes can be read off from CLEAR \textit{Image}’s regression equation. Using the example in Figure \ref{CLEAR_over} with the logistic formula, a classification probability of $>$ 0.99 requires $\textbf{\textit{w}}^T\textit{\textbf{x }}  > 4.6$.  The GAN healthy image corresponds to all the binary segment variables being equal to 0. Hence, $\textbf{\textit{w}}^T\textit{\textbf{x }}$ is equal to the intercept value of -4.9, giving a probability of \( (1+exp^{4.9})^{-1} \approx 0.01\). If a segment $s_i'$ is now replaced by $s_i$, the corresponding binary variable changes to 1. Hence if segment 9 is infilled, then \textit{Seg09} = 1 and $\textbf{\textit{w}}^T\textit{\textbf{x }} =6.8 \ (i.e. 11.7 - 4.9)$. Similarly, infilling just segment 11 will make $\textbf{\textit{w}}^T\textit{\textbf{x }} > 4.6$. Either substitution is sufficient to guarantee  $\textbf{\textit{w}}^T\textit{\textbf{x }} > 4.6$, irrespective of any other changes that could be made to the values of the other segment variables. Hence segments 9 and 11 are each a sufficient cause leading to overdetermination.

By contrast, XAI methods such as LIME and Kernel SHAP cannot identify cases of overdetermination. This is because they use simple linear regression instead of logistic regression. For example, suppose that an image has three segments: $s_1, s_2, s_3$. In the regression dataset, each segment infilled from $x$ has a value of 1 and each segment infilled from $x'$ has a value of 0. LIME/Kernel SHAP’s regression equation will have the form: $y = k_1s_1 +k_2s_2 + k_3s_3$ . In the case of LIME, $y$ is meant to be the classification probability and the regression coefficients ($k_1, k_2, k_3$) are the feature importance scores.  Let us suppose there is overdetermination, with segments $s_1$ and $s_2$ each being a sufficient cause for $x$ to be in a given class (e.g. 'pleural effusion') with more than 0.99 probability. Hence, the regression equation should set $y$ to a value greater than 0.99 not only when $s_1 = s_2$ = 1, but also when either $s_1 =1$ or $s_2 =1$. This is clearly impossible with the above linear form (and the constraint that $y \leq 1$). Mutatis mutanda, the same argument applies for Kernel SHAP.

\section{Experimental Investigation}

There are two case studies, the first using a synthetic dataset, the second analysing pleural effusion X-rays taken from the CheXpert dataset \cite{irvin2019chexpert}. Transfer learning was used to train both a VGG-16 with batch normalisation and a DenseNet-121 classifier for each dataset. CLEAR \textit{Image} was evaluated against Grad-CAM, Extremal Perturbations and LIME. The evaluation consisted of both a qualitative comparison of saliency maps and a comparison of pointing game and intersection over union (IoU) scores. CLEAR \textit{Image}’s fidelity errors were also analysed (none of the other XAI methods measures fidelity).

\subsection{Datasets}
\label{sec:datasets}
The synthetic dataset's images share some key characteristics found in medical imaging including: (i) different combinations of features leading to the same classification (ii) irrelevant features. All images (healthy and diseased) contain a set of concentric circles, a large and a small ellipse. An image is ‘diseased’ if either: (1) the small ellipse is thin-lined, and the large ellipse contains a square or (2) there is a triangle, and the large ellipse contains a square. The dataset is an adaptation of \cite{wolleb2020descargan}.

CheXpert is a dataset of chest X-rays with automated pathological label extraction through radiology reports, consisting of 224,316 radiographs of 65,240 patients in total. Images were extracted just for the classes ‘pleural effusion’ and ‘no finding’. Mis-classified images and images significantly obstructed by supporting devices were manually filtered. A random frontal X-ray image per patient was collected. In total, a dataset of 2,440 images was used in this work for model training, validation and testing. Appendix A.2 details the data preparation process. A hospital doctor provided the ground truth annotation to the X-ray images with pleural effusion for our case study. 


\subsection{Evaluation Metrics}

This paper uses two complementary metrics to evaluate XAI methods. Both require annotated images identifying ‘target’ regions that should be critical to their classification. A pointing game produces the first metric, which measures how successfully a saliency map ‘hits’ an image’s targets. Previously pointing games have been designed for cases where (i) images have single targets (ii) the saliency maps have a maximum intensity point \cite{fong2019understanding,zhang2018top}. By contrast, this paper’s case studies have multiple targets, and the pixels within each CLEAR \textit{Image} segment have the same value. We, therefore, formulated a novel pointing game. The pointing game partitions a ‘diseased’ image into 49 square segments, P = \{$p_1\ldots p_{49}$\} and identifies which squares contain each of the targets. The corresponding saliency map is also partitioned, and each square is allocated a score equal to the average intensity of that square’s pixels Q = \{$q_1 \ldots q_{49}$\}. The pointing game then starts with the $q_i$ of highest intensity and determines if the corresponding $p_i$ contains a relevant feature. A successful match is a ‘hit’ and an unsuccessful match is a ‘miss’. This process continues until every target has at least one hit. The score for an image is the number of hits over the number of hits plus misses. Pseudocode is provided in Algorithm \ref{pointing_algo}.

The second metric is IoU. It is assumed that each pixel in a saliency map is classified as ‘salient’ if it is above $70^{th}$ percent of the maximum intensities in that map.  IoU then measures the overlap between the ‘salient’ pixels $pix^{salient}$  and the pixels belonging to the image’s targets $pix^{target}$ : $ IOU =  \bigl( pix^{salient} \ \cap \ pix^{target} \bigr) / \bigl( pix^{salient} \cup \ pix^{target} \bigr)$. The chosen percentile was an empirically identified threshold to maintain a relatively high IoU score by balancing high intersection with $pix^{target}$ and small union of pixel regions with a large enough $pix^{salient}$ (see Appendix \ref{apx:sup_results} for details).

Both metrics are useful but have counterexamples. For example, IoU would give too high a score to a saliency map that strongly overlapped with a large image target but completely missed several smaller targets that were also important to a classification. However, applied together, the two metrics provide a good indication of an XAI’s performance.

\begin{figure}[!htbp]
  \centering
    \begin{minipage}[h]{.7\textwidth}
    \begin{algorithm}[H]
    \SetKwInOut{Input}{input}\SetKwInOut{Output}{output}
    \SetAlgoLined
    \DontPrintSemicolon
    \caption{Pointing Game}
    \Input{ $x$ - input image, $A$ - annotated features\\ $w$ - XAI saliency map}
    $P_A\leftarrow\ \;$Square\_Idx\_of\_Each\_Feature({$A,x$})\\ 
    $Q\leftarrow\ \;$Average\_Intensity\_Each\_Square({$w$})\\ 
    $Q'\leftarrow\ \;$Square\_Idx\_Sort\_Highest\_Intensity({$Q$})\\
    $hits\leftarrow\ 0; \;misses\leftarrow\ 0;h_A\leftarrow\ False$ \\
    \ForEach(\newline \tcp*[h]{starting with highest}){$q'_i \in Q'$}{
    \ForEach{$a_j \in A$}{
    \uIf(\tcp*[h]{square idx match}){$q'_i \in P_{a_j}$}{
    $hits\leftarrow\ \;hits + 1\; ; h_{a_j}\leftarrow\ \;True$\;}
    \uElse{
    $misses\leftarrow\ \;misses + 1$\;}
     }
    \uIf(\newline \tcp*[h]{exit once all features hit}){$\forall a_j \; (h_{a_j} = True)$}{\textbf{break}}
     }
    \Return  $<hits, misses>$
     \label{pointing_algo}
    \end{algorithm}
    \end{minipage}
\end{figure}

\subsection{Experimental Runs}
CLEAR  \textit{Image} was run using logistic regression with the Akaike information criterion; full testing and parameter values can be found in Appendix B.3. The test datasets consisted of 95 annotated X-rays and 100 synthetic images. The average running time for CLEAR  \textit{Image} was 20 seconds per image for the synthetic dataset and 38 seconds per image for the CheXpert dataset, running on a Windows i7-8700 RTX 2070 PC. Default parameter values were used for the other XAI methods, except for the following beneficial changes: Extremal Perturbations was run with ‘fade to black’ perturbation type, and using areas \{0.025,0.05,0.1,0.2\} with the masks summed and a Gaussian filter applied. LIME was run using Quickshift segmentation with kernel sizes 4 and 20 for the CheXpert and synthetic datasets respectively. 

\subsection{Experimental Results}
\label{sec:expt_result}
\begin{figure}[!htbp]
  \includegraphics[width=0.95\textwidth]{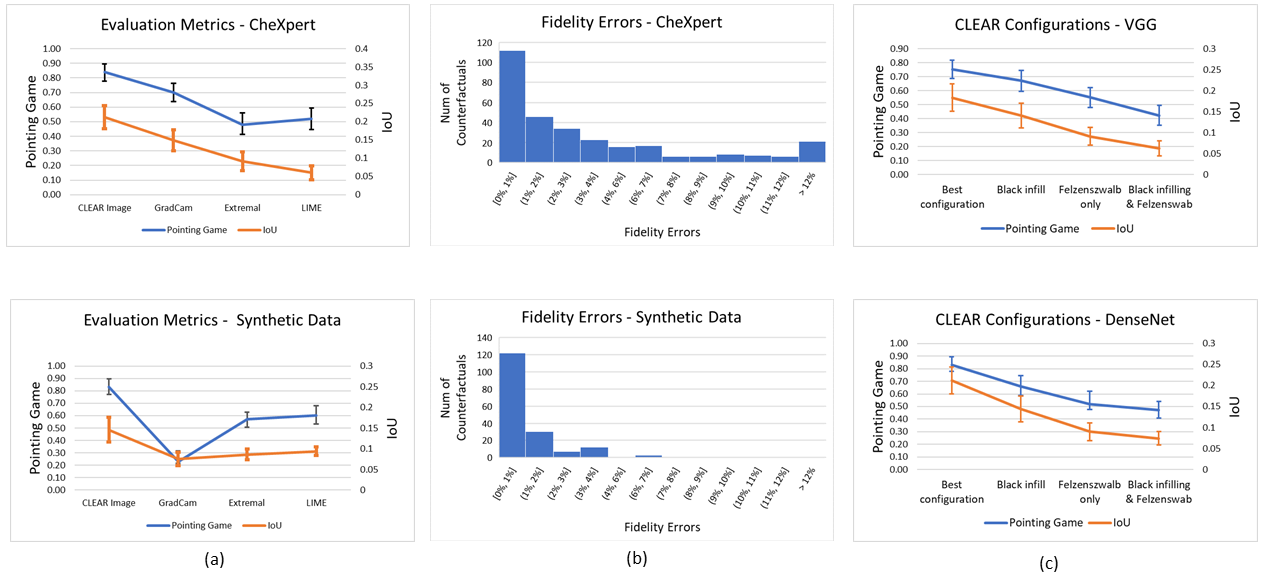}  
  \centering
  \caption{Evaluation metrics. Figure (a) compares the performances of different XAI methods with the DenseNet models. Figure (b) shows the fidelity errors for the DenseNet models. Figure (c) compares the performances of different configurations of CLEAR  \textit{Image}. The bars show 95\% confidence intervals.}
  \Description{Quantitative evaluation results using pointing game and IoU metrics}
  \label{CLEAR_Results}
\end{figure}

\begin{figure}[!htbp]
  \includegraphics[width=0.8\textwidth]{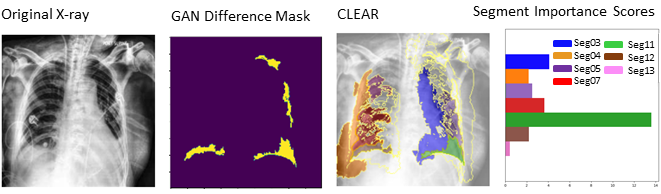}  
  \centering
  \caption{GAN-Augmented Segmentation versus GAN difference mask. The difference mask identifies four segments but when CLEAR \textit{Image} perturbs these, the two nearest to the top were found to be irrelevant. Of the other two segments, CLEAR \textit{Image} identifies the segment it colors green to be far more important to the classification probability.}
  \Description{Comparison of GAN-Augmented Segmentation and GAN difference mask for segment generation.}
  \label{Augmented_GAN}
\end{figure}

CLEAR \textit{Image} outperforms the other XAI methods on both datasets (Figure \ref{CLEAR_Results}a). Furthermore, its fidelity errors are low, indicating that the regression coefficients are accurate for the counterfactually important segments (Figure \ref{CLEAR_Results}b). Figure \ref{CLEAR_Results}c illustrates some of the benefits of using the 'Best Configuration', which uses GAN-augmented segmentation and infills using $x'$. This is compared with (i) segmenting with Felzenszwalb and infilling with $x'$ (ii) segmenting with GAN-augmented but infilling with black patches (iii) segmenting with Felzenszwalb, infilling with black patches.  Figure \ref{Augmented_GAN} illustrates how CLEAR \textit{Image}'s use of GAN-augmented leads to a better explanation than just using a difference mask (e.g. CLEAR \textit{Image}’s performance was similar for VGG-16 and DenseNet; therefore, only the DenseNet results are presented unless otherwise stated.

\begin{figure}[!hp]
  \includegraphics[scale=0.58]{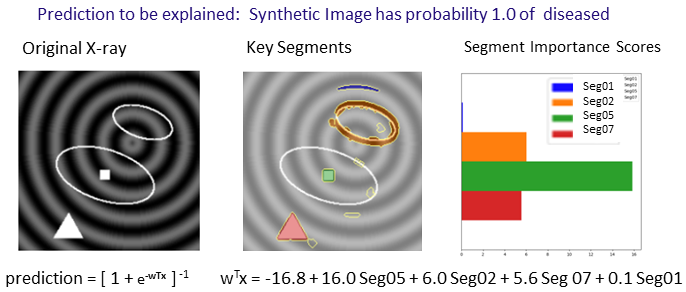}  
  \centering
  \caption{Extracts from a CLEAR \textit{Image} report for a synthetic image.  The regression equation shows that Seg05 is a \textit{necessary but insufficient cause} of the X-ray being diseased.}
  \Description{An extract from a CLEAR $Image$ report for a synthetic image.}
  \label{CLEAR_synthetic_report}
\end{figure}

\begin{figure}[!htbp]
  \includegraphics[width=0.8\textwidth]{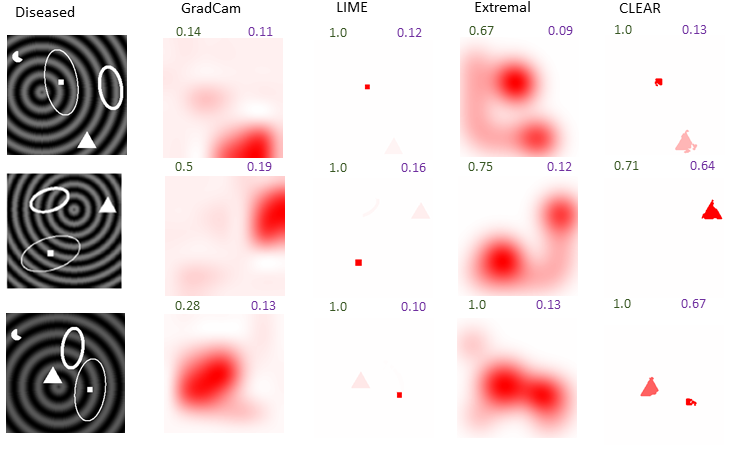}
  \centering
  \caption{Comparison of XAI methods on synthetic data. The pointing game scores are shown in green and the IoU scores are in purple. The maps illustrate how CLEAR \textit{Image} and LIME are able to tightly focus on salient regions of an image compared to broadbrush methods such as Grad-CAM and Extremal. The significance of a patch is indicated by its red intensity.}
  \Description{Comparison of XAI methods on synthetic data with corresponding pointing game and IoU scores.}
  \label{synthetic_images}
\end{figure}

CLEAR \textit{Image}’s regression equation was able to capture the relatively complex causal structure that generated the synthetic dataset. Figure \ref{CLEAR_synthetic_report} shows an example. A square (SQ) is a \textit{necessary but insufficient} cause for being diseased. An image is labelled as diseased if there is also either a triangle (TR) or the small ellipse is thin-lined (TE). When SQ, TR and TE are all present in a single image, there is a type of overdetermination in which TR and TE are each a sufficient cause \textit{relative} to the ‘image with SQ already present’. As before, a diseased image corresponds to the binary segment variables equalling one and a classification probability of being diseased  $>0.99$ requires $\textbf{\textit{w}}^T\textit{\textbf{x }}  > 4.6$. This can only be achieved by Seg 5 (corresponding to SQ) plus at least one of Seg 2 or Seg 7 (TE, TR) being set to 1 (i.e. being present).  Figure \ref{synthetic_images} compares the saliency maps for synthetic data.

For the CheXpert dataset, figure \ref{X-ray_images} illustrates how CLEAR \textit{Image} allows for a greater appreciation of the pathology compared to `broad-brush' methods such as Grad-CAM (please see Appendix A1 for further saliency maps). Nevertheless, the IoU scores highlight that the segmentation can be further improved. For CheXpert's counterfactuals, only 5\% of images did not have a counterfactual with four or fewer $s'$ segments. Most images required several $s$ segments to be infilled before its classification flipped to ‘healthy’, 17\% required one segment, 30\% with two segments, 24\% with three segments and 24\% with four segments. 17\% of the X-rays' were found to be causally overdetermined.

\section{Conclusion and Future Work}

A key reason for CLEAR \textit{Image}'s outperformance of other XAI methods is its novel use of GANs. It recognises that a difference mask is only the starting point for an explanation. Instead, it uses a GAN image both for infilling and as an input into its own segmentation algorithm.

\begin{figure}[!htbp]
  \includegraphics[width=0.8\textwidth]{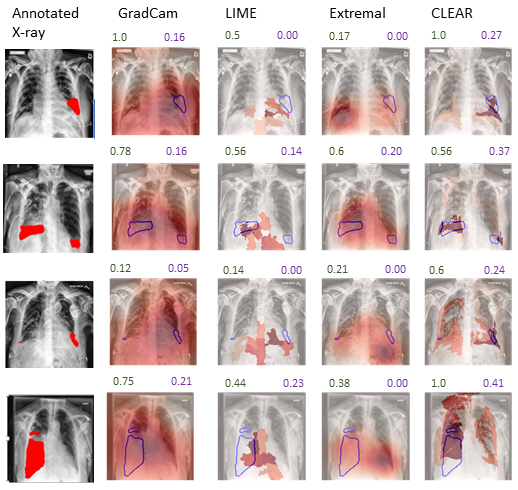}  
  \centering
  \caption{Comparison of XAI methods on X-ray. The pointing game scores are shown in green and the IoU scores are in purple. The significance of a patch is indicated by the intensity of red against the blue outlined annotated ground truth.}
  \Description{Comparison of XAI methods on X-ray image data with corresponding pointing game and IoU scores.}
  \label{X-ray_images}
\end{figure}

As AI systems for image data are increasingly adopted in society, understanding their implicit causal structures has become paramount. Yet the explanations provided by XAI methods cannot always be trusted, as the differences in Figure \ref{X-ray_images}’s saliency maps show. It is therefore critical that an XAI method measures its own fidelity. By ‘knowing when it does not know’, it can alert the user when its explanations are unfaithful.

The examples in this paper help to illustrate our claim that XAI counterfactual methods will often fail to provide satisfactory explanations of a classifier's local input-output behaviour. This is because a satisfactory explanation requires both counterfactuals and a supporting causal equation. It is only because CLEAR \textit{Image} produces a causal equation that it is then able to identify (a) segment importance scores, including identifying segments with negative scores (Figure \ref{CLEAR_report}) (b) segments that are necessary but insufficient causes (Figure \ref{CLEAR_synthetic_report}) (c) cases of overdetermination (Figure \ref{CLEAR_over}). Providing only counterfactuals by themselves is clearly insufficient; imagine another science, say physics, treating a statement of counterfactuals as being an explanation, rather than seeking to discover the governing equation. Perhaps the primary benefit of XAI counterfactual methods is in suggesting sets of actions, but as Karimi et al. \cite{karimi2021algorithmic} argue, counterfactual methods will often perform suboptimally at this task.

A limitation of CLEAR \textit{Image} is that it first requires training a GAN, which can be a challenging process. Another possible limitation could be the understandability of  CLEAR \textit{Image} to non-technical users. However, its reports can be suitably tailored, e.g. only showing saliency maps, lists of counterfactuals and cases of overdetermination.
 
We have shown that CLEAR \textit{Image} can illuminate cases of causal overdetermination. Many other types of causal structures may also be ubiquitous in AI. For example, causal pre-emption and causal clustering are well documented within the philosophy of science \cite{baumgartner2009inferring,schaffer2004trumping}. The relevance of these to XAI will be a future area of work. A user study should also be carried out. However, these are time/resource consuming and need to be devised carefully by experts within specific application domains to produce sound, reliable results. Instead, we focus on objective measures and evaluations of XAI research which in our view must precede any user study. Future work will also focus on improving segmentation, e.g. by introducing domain-specific constraint parameters for GANs, to minimise the modifications of specified attributes (e.g. changes in the heart when generating lung X-rays).

\clearpage
\printbibliography


\clearpage
\appendix

\section{Supplemental results for CheXpert dataset and associated data pre-processing}
\label{adx:data_prep}

\subsection{Supplementary qualitative results} \label{apx:sup_results}

\begin{figure*}[htbp]
    \centering
    \subfloat{
        \label{fig:iou_cxp_d2}
        \includegraphics[width=0.7\textwidth]{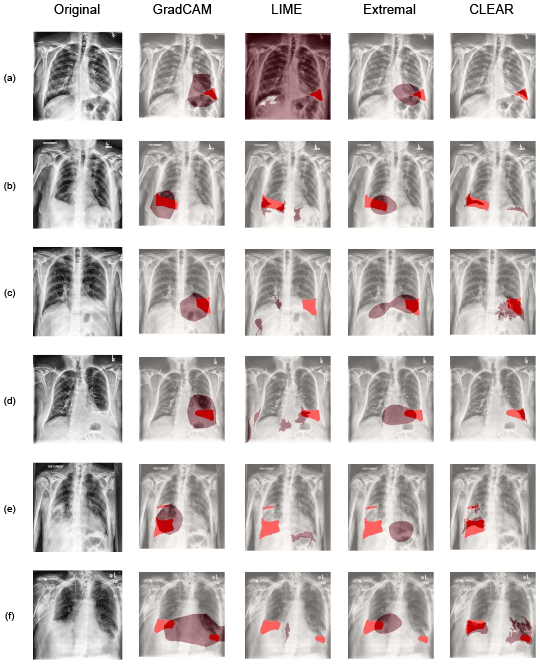}
    }
    \caption{Representative comparative examples of the identified important segments of a DenseNet-based image classification model (Val Acc: 98.8\%) for pleural effusion using (i) CLEAR \textit{Image}, (ii) Grad-CAM, (iii) Extremal Perturbation and (iv) LIME.}%
    \Description{Comparative examples of identified important segments of a DenseNet-based image classification model across XAI methods.}
    \label{ref_iou_cxp_d}
\end{figure*}

\begin{figure*}[htbp]
    \centering
    \subfloat{
        \label{fig:iou_cxp_v2}
        \includegraphics[width=0.7\textwidth]{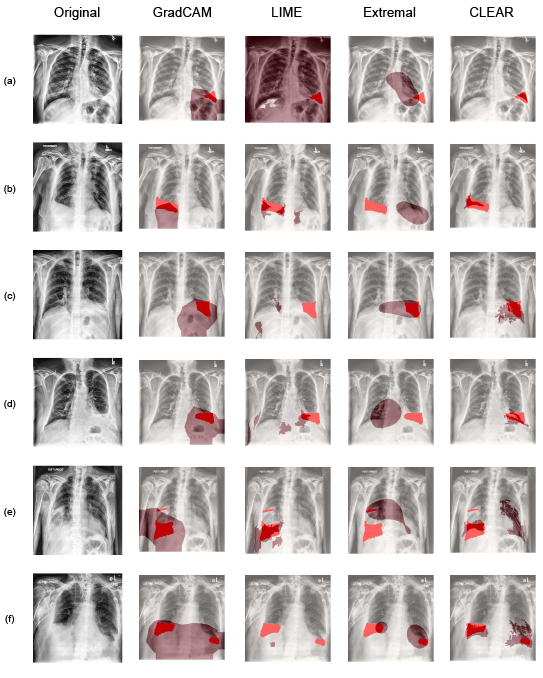}
    }
    \caption{Representative comparative examples of the identified important segments of a VGG16-based image classification model (Val Acc: 97.5\%) for pleural effusion using (i) CLEAR \textit{Image}, (ii) Grad-CAM, (iii) Extremal Perturbation and (iv) LIME.}%
     \Description{Comparative examples of identified important segments of a VGG16-based image classification model across XAI methods.}
    \label{ref_iou_cxp_v}
\end{figure*}

\begin{figure}[!htbp]
  \centering
  \subfloat[Comparison of XAI methods on VGG16 Model]{\includegraphics[width=0.45\columnwidth]{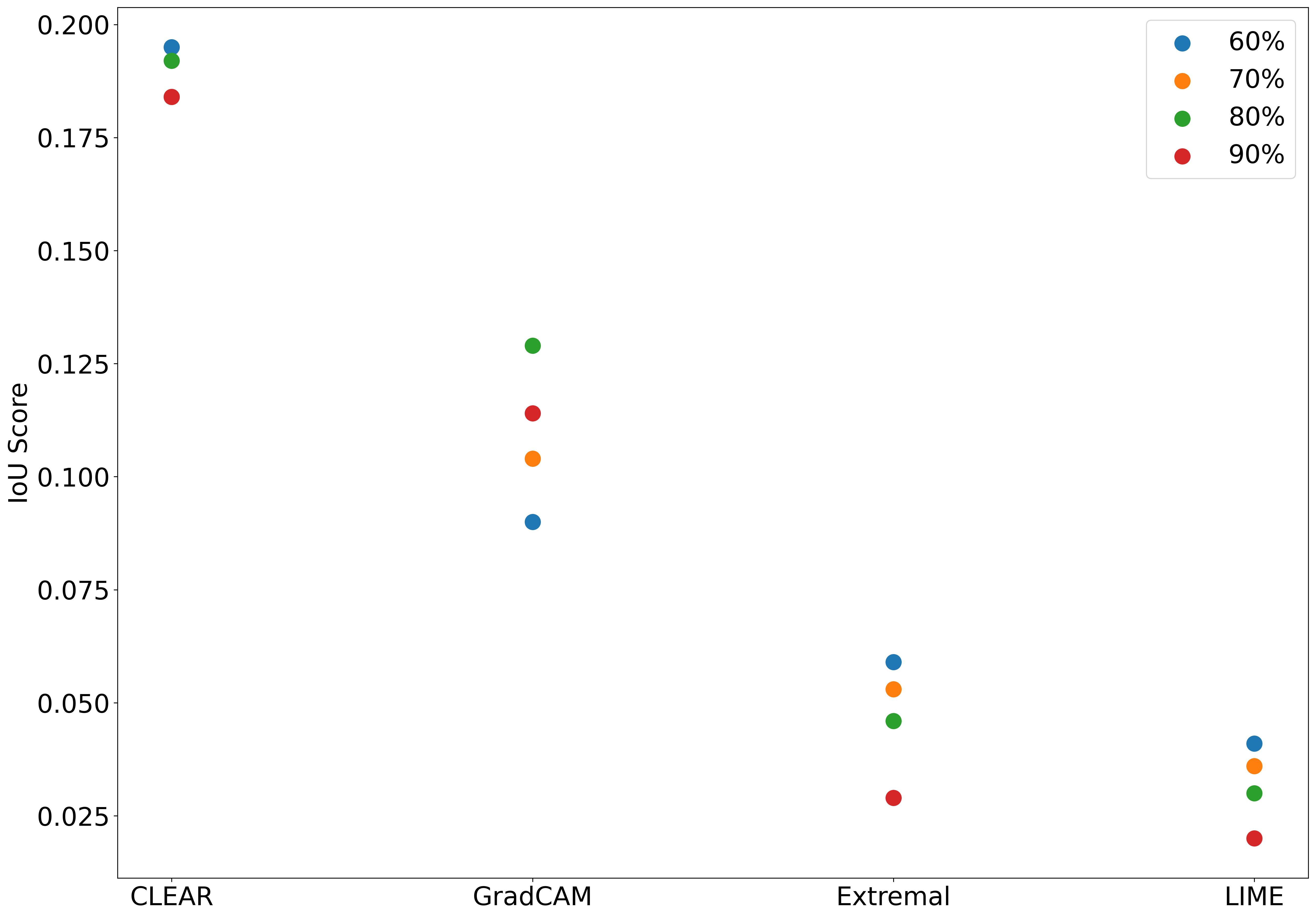} \label{fig:vgg_iou_thresh}} \\
  \subfloat[Comparison of XAI methods on DenseNet Model]{\includegraphics[width=0.45\columnwidth]{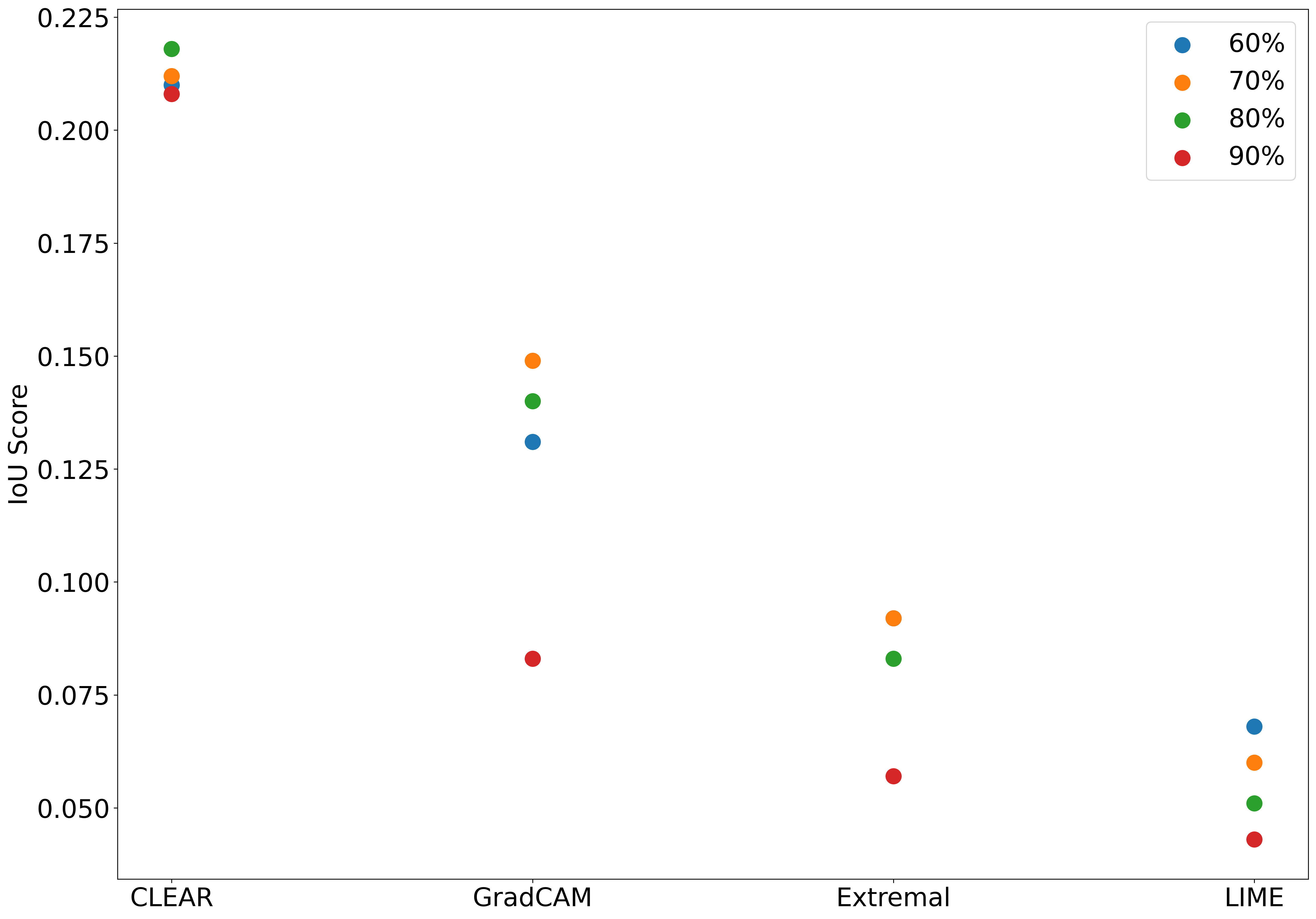} \label{fig:densenet_iou_thresh}}\\
  \caption{Comparison of IoU score against four XAI methods, (1) CLEAR \textit{Image}, (2) GradCAM, (3) Extremal and (4) LIME to determine the threshold of intensity at 10\% intervals. CLEAR \textit{Image} outperforms the other XAI methods for each of the 4 intensity thresholds.} \label{fig:thresh_iou_evaluate}
  \Description{Evaluation of intensity threshold over XAI methods as a hyperparameter used in IoU metrics at 10\% intervals.}
\end{figure}

Additional qualitative results for the CheXpert dataset are presented in this section (Figures \ref{ref_iou_cxp_d} and \ref{ref_iou_cxp_v}) where the most important segments (regions) identified by each XAI method is matched against the annotated ground truth. These are the pixels of saliency maps that are above 70 percent of the maximum intensity (i.e. the segments used to calculate the IoU scores). This threshold was determined empirically to yield high IoU score across all the XAI methods evaluated (see Figure \ref{fig:thresh_iou_evaluate}).

Figure \ref{ref_iou_cxp_d} shows the additional results for the DenseNet model while Figure \ref{ref_iou_cxp_v} presents the results for the VGG16 model. These results have demonstrated higher precision using CLEAR Image in identifying significant segment matching against the annotated ground truth in comparison to other explanation methods. These two figures provide a qualitative comparison to supplement the results presented in Figure \ref{CLEAR_Results} where CLEAR \textit{Image} outperforms other XAI methods.

\clearpage
\subsection{Data Pre-Processing}
CheXpert has a total of 14 pathological classes including ‘No Finding’, and these are labelled through an automated rule-based labeller from text radiology reports. For each observation, the Stanford team has classified each radiograph as either negative (0), uncertain (-1) or positive (1). Other metadata includes gender, age, X-ray image projection and presence of supporting devices.

In this study, this dataset (v1.0) was applied for the model development of a binary classification task to demonstrate the capability of CLEAR \textit{Image} as an XAI framework. An initial filtering process of the metadata was applied for the two classes used in the study - (1) Diseased with Pleural Effusion and (2) Healthy (this was assumed to be X-ray images with no findings and no positive observations in any of the pathological conditions). To minimise potential complications with other pathological conditions, X-ray images with only positive in pleural effusion were used with  the other pathological categories either as negative/blank.

\begin{figure}[!htb]%
    \centering
    \subfloat[\centering Diseased]{{\includegraphics[width=0.45\textwidth]{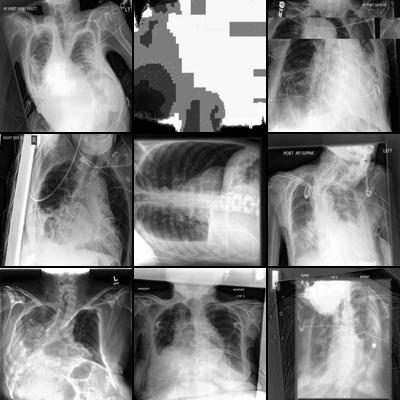} }}
    \qquad
    \subfloat[\centering Healthy]{{\includegraphics[width=0.45\textwidth]{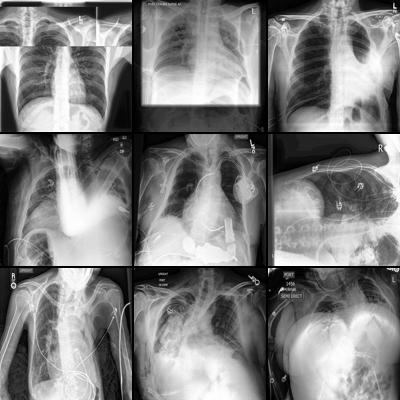} }}%
    \caption{Representative examples of poorly curated images including image distortion, mis-orientation, obstruction by limbs and support devices as well as significant spine deformation.}%
    \Description{Representative examples of poorly curated X-ray images.}
    \label{fig:xpert_poor}%
\end{figure}

A review of the filtered images also identified that the dataset was curated with some images having significant artefacts that can hamper model training performance. Figure \ref{fig:xpert_poor} presents some of these images in both diseased and healthy categories. Many of these consisted of artefacts from image capturing and processing (e.g. image distortion, orientation, low resolutions or miscalibration). Some images were also significantly obstructed by limbs or support devices. Some healthy images were also wrongly labelled according to a hospital doctor, who assisted in our project. A secondary manual filtering was conducted to remove any identified images with artefacts.

\begin{figure}[!htb]%
    \centering
    \subfloat[\centering Diseased]{{\includegraphics[width=0.45\textwidth]{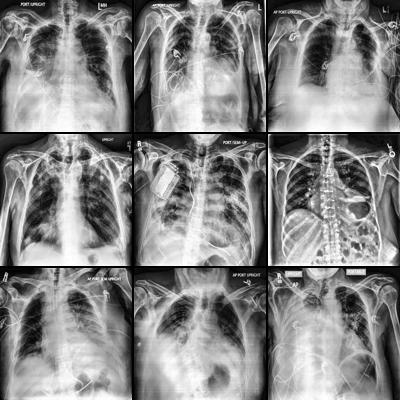} }}%
    \qquad
    \subfloat[\centering Healthy]{{\includegraphics[width=0.45\textwidth]{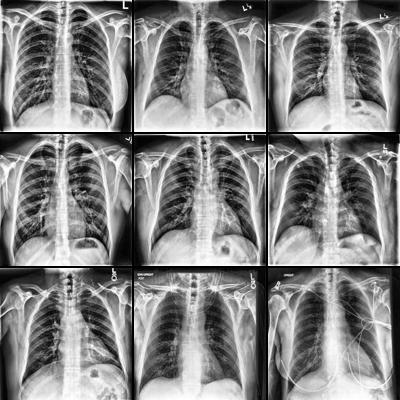} }}%
    \caption{Representative examples of final images for (a) diseased with identifiable regions of pathology and (b) healthy images with clear air space. All images have minimal obstructions from support devices.}%
    \Description{Representative examples of final X-ray images used in this work.}
    \label{fig:xpert_good}%
\end{figure}

The 2440 selected images were split approximately 80/10/10 for the training/validation/testing. The images were also resized to 256 x 256 as the input into the classification model and generative adversarial network (GAN) as described in Section 5. Figure \ref{fig:xpert_good} presents some typical images in the final dataset for both diseased and healthy categories.

\clearpage
\section{Model Parameters}
\label{adx:gan}

\subsection{DeScarGAN and Parameters}

The DeScarGAN architecture was adopted for the synthetic dataset in Section 5.1. 80\% of the dataset (4000 images) was used for GAN training and 20\% of the dataset (1000 images) was used for validation. A total of 2,500 epochs was run and the best epoch was selected on visual quality. Additional 100 images were generated as an out-of-sample test dataset. Adam optimizer was used with $\beta_1$ = 0.5, $\beta_2$=0.999. An initial learning rate of $10^{-4}$ was used and stepped down to a final learning rate of $10^{-6}$. Default hyperparameters for loss functions were used to mimic a similar investigation from the original author as shown below in Table \ref{tab:1}:

\begin{table}[ht]
  \centering
  \begin{tabular}{ll}
    \textbf{Loss Term} & \textbf{Weight Value}  \\
    Adversarial Loss: & $\lambda_{adv,g}$ = 1 (Generator) \\
     & $\lambda_{adv,d}$ = 20 (Discriminator) \\
    Gradient Penalty Loss: &	$\lambda_{gp}$ = 10 \\
    Identity Loss: 	  &		$\lambda_{id}$ = 50 \\
    Reconstruction Loss:  &	$\lambda_{rec}$ = 50 \\
    Classification Loss: &	$\lambda_{cls,g}$ = 1 (Generator) \\ & $\lambda_{cls,d}$ = 5 (Discriminator)\\

  \end{tabular}
  \caption{Default Loss Function Hyperparameters used in DeScarGAN}
  \label{tab:1}
\end{table}

\subsection{StarGAN-V2 and Parameters}

StarGAN-V2 \cite{choi2020starganv2} has been adopted in this work as a state-of-art GAN network for image translation. The GAN provided the necessary contrastive images for the CheXpert dataset. Default hyperparameters were maintained while notable loss weights are highlighted in Table \ref{tab:stargan}. Adam optimizer was used with $\beta_1$ = 0, $\beta_2$=0.99. A total of 50,000 epochs were run for the CheXpert dataset. The style encoding was referenced to the input image for the translation to the targeted class. This aided in maintaining the general features of the images compared to the original. As StarGAN-V2 \cite{choi2020starganv2} did not constrain its generation to a localised region (e.g. lungs), post-processing of segmentation and blending was implemented for the CheXpert dataset. Segmentation of the lung region was based on a pre-trained model with a U-Net architecture. The segmentation mask was subsequently used to guide the replacement of pixels within the lung region from the GAN generated healthy image onto the original diseased image. Gaussian Blur was applied to minimise the edge effect during the blending process. This post-processing step aided in restricting the feature identification space within the lungs and reducing the computational cost for locating the counterfactuals.

\begin{table}[h]
  \centering
  \begin{tabular}{ll}
    \textbf{Loss Term} & \textbf{Weight Value}  \\
    Style Reconstruction Loss: & $\lambda_{sty}$ = 1 \\
    Style Diversification Loss: &	$\lambda_{ds}$ = 1 \\
    Cyclic Loss:  &	$\lambda_{cyc}$ = 1 \\

  \end{tabular}
  \caption{Default Loss Function Hyperparameters used in StarGANv2 \cite{choi2020starganv2}}
  \label{tab:stargan}
\end{table}

An evaluation of similarity to real healthy images was performed using the Fréchet inception distance (FID) \cite{heusel_fid} benchmarking against the set of healthy images in the model training dataset. Four image sets were compared: (1) real healthy images in the validation set, set of images with pleural effusion processed as described in Figure \ref{postGANpipeline} with replacement of lung segments using (2) corresponding GAN-generated healthy images, (3) Gaussian blurred version of the original images and (4) constant value of zero (i.e. black). This FID score indicated how close each of the four compared image sets to the benchmark images in the training set. A low score indicated similarity between the two datasets.

\begin{figure}[!htbp]
  \includegraphics[width=0.5\textwidth]{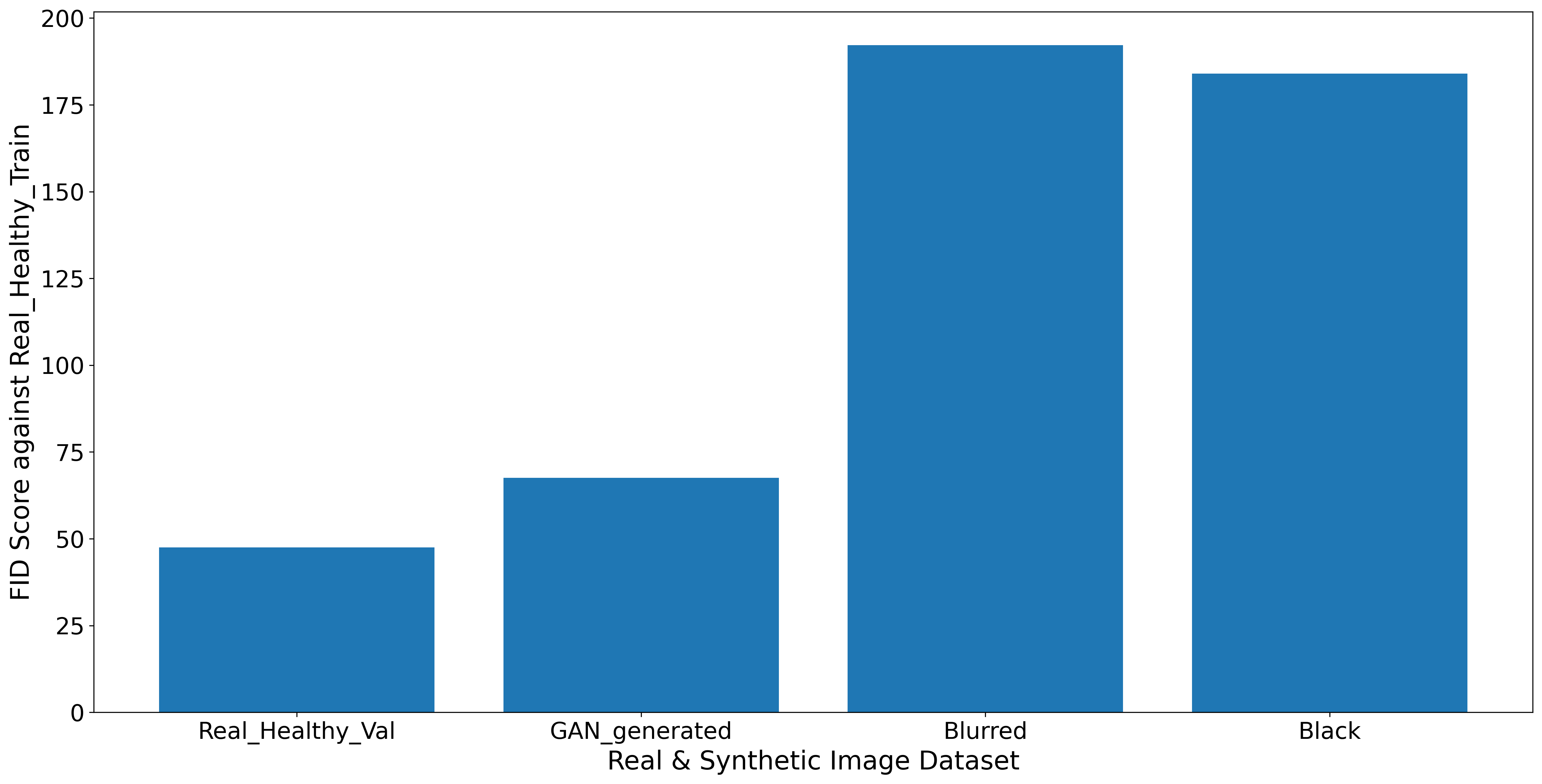}  
  \centering
  \caption{Comparison of Fréchet inception distance (FID) against the training healthy image dataset with (1) a set of real healthy images in the validation set, set of images with pleural effusion processed as described in Figure \ref{postGANpipeline} with replacement of lung segments using (2) corresponding GAN-generated healthy images, (3) Gaussian blurred version of the original images and (4) constant value of zero (i.e. black).}
  \Description{Similarity evaluation against training healthy images with lung segment replacement for X-ray image using GAN-generated healthy image, Gaussian blurred original image and constant value of zero (i.e. black) using Fréchet inception distance (FID).}
  \label{FID_compare}
\end{figure}

As observed in Figure \ref{FID_compare}, the processed images with replacement using GAN generated healthy lung segments resemble more similar to actual healthy images than blurred or black segments. As such, GAN generated processed images as described in Figure \ref{postGANpipeline} were selected as the choice of synthetic healthy images for this work.

\clearpage
\subsection{Clear \textit{Image} Parameters}

The default parameters used for the Chest X-ray experiments were:
\begin{verbatim}
case_study = 'Medical' 
max_predictors = 6
num_samples = 1000
regression_type = 'logistic'
logistic_regularise = False
score_type = 'aic'
apply_counterfactual_weights = True
counterfactual_weight = 200
binary_decision_boundary = 0.5
no_polynomimals_no_interactions = True
interactions_only = True
no_intercept = False
centering = True
include_features = False
include_features_list = []
sufficiency_threshold = 0.99
image_infill ='GAN'
image_all_segments= False
threshold_method = 'manual'
image_use_old_synthetic = False
image_counterfactual_interactions = False
image_segment_type ='Augmented_GAN'
max_segments_in_counterfactual =4
min_segs_created_for_Augmented_GAN = 4
min_seg_size = 250
min_seg_increment = 25
image_classes =['normal','effusion']
\end{verbatim}
\noindent
\newline
For the CLEAR \textit{Image} configuration experiments the parameter 'image\_infill' had values ['GAN', 'black'] and the parameter image\_segment\_type had values ['Augmented\_GAN', 'Felzenszwalb']\\

The same parameter values were used for the synthetic case study except:
\begin{verbatim}
case_study = 'Synthetic'
image_segment_type ='Thresholding'
\end{verbatim}

\end{document}